\documentclass[10pt,twocolumn,letterpaper]{article}

\usepackage{iccv}
\usepackage{times}
\usepackage{epsfig}
\usepackage{graphicx,booktabs,array}
\usepackage{amsmath}
\usepackage{amssymb}

\usepackage[pagebackref=true,breaklinks=true,letterpaper=true,colorlinks,bookmarks=false]{hyperref}

\iccvfinalcopy % *** Uncomment this line for the final submission

\def\iccvPaperID{7909} % *** Enter the ICCV Paper ID here

\ificcvfinal\fi

\graphicspath{ {./figures/}}

\begin{document}

%%%%%%%%% TITLE
\title{kNet: A Deep kNN Network To Handle Label Noise}

\author{Itzik Mizrahi\\
Tel Aviv University\\
{\tt\small itzikmizrahi@mail.tau.ac.il}
\and
Shai Avidan\\
Tel Aviv University\\
{\tt\small avidan@eng.tau.ac.il}
}

\maketitle
\ificcvfinal\thispagestyle{empty}\fi

%%%%%%%%% ABSTRACT
\begin{abstract}
    Deep Neural Networks require large amounts of labeled data for their training.
    Collecting this data at scale inevitably causes label noise. Hence, the need to develop learning algorithms that are robust to label noise. 
    
    In recent years, k Nearest Neighbors (kNN) emerged as a viable solution to this problem. Despite its success, kNN is not without its problems. Mainly, it requires a huge memory footprint to store all the training samples and it needs an advanced data structure to allow for fast retrieval of the relevant examples, given a query sample.
    
    We propose a neural network, termed kNet, that learns to perform kNN. Once trained, we no longer need to store the training data, and processing a query sample is a simple matter of inference. To use kNet, we first train a preliminary network on the data set, and then train kNet on the penultimate layer of the preliminary network.
    
    We find that kNet gives a smooth approximation of kNN, and cannot handle the sharp label changes between samples that kNN can exhibit. This indicates that currently kNet is best suited to approximate kNN with a fairly large $k$. Experiments on two data sets show that this is the regime in which kNN works best, and can therefore be replaced by kNet. In practice, kNet consistently improve the results of all preliminary networks, in all label noise regimes, by up to $3\%$.
    
    % We encourage applying kNet to any (noise-robust or standard) trained network under noisy environment, as a post fine learning stage. Our experiments shows that consistently it will increase the accuracy ,empirically up to 3\%.
    
\end{abstract}
%%%%%%%%% BODY TEXT

\section{Introduction}
Training deep neural networks (DNN) usually requires large amounts of labeled data. Obtaining this data is challenging, especially where expert annotators are involved.
And even expert annotators are prone to make errors. Alternatively, one can use non-expert annotators, as well as web harvesting to obtain large amounts of data with noisy labels cheaply. Both cases lead to the realisation that machine learning algorithms must be able to handle label noise.

Some report that neural networks are resilient to label noise, but this depends on the type of label noise. The easiest label noise to handle is the uniform noise, where the label of a sample is flipped with uniform probability to some other label. A much more challenging label noise to deal with is the one where two labels are confused, for example, two breeds of dogs that are visually similar.

In recent years, $k$ Nearest Neighbor (kNN) classification emerged as a viable option to handle learning with label noise. Theoretical \cite{DBLP:journals/corr/GaoNZ16}, as well as empirical \cite{DBLP:conf/icitcs/AgrawalR15} results show that, for different types of label noise, kNN is robust enough to classify most data correctly. Moreover \cite{DroryRAG20} show correlation between a DNN's prediction to the labels of the training examples in its local neighborhood.

On the downside, kNN must store all the training samples in memory which leads to a large memory foot print. In addition, kNN algorithms often require the use of approximate nearest neighbor data structures to support fast retrieval. 

\begin{figure}[t!]
\begin{center}
\begin{tabular}{l}
\includegraphics[width=8.6cm,height=4.3cm]{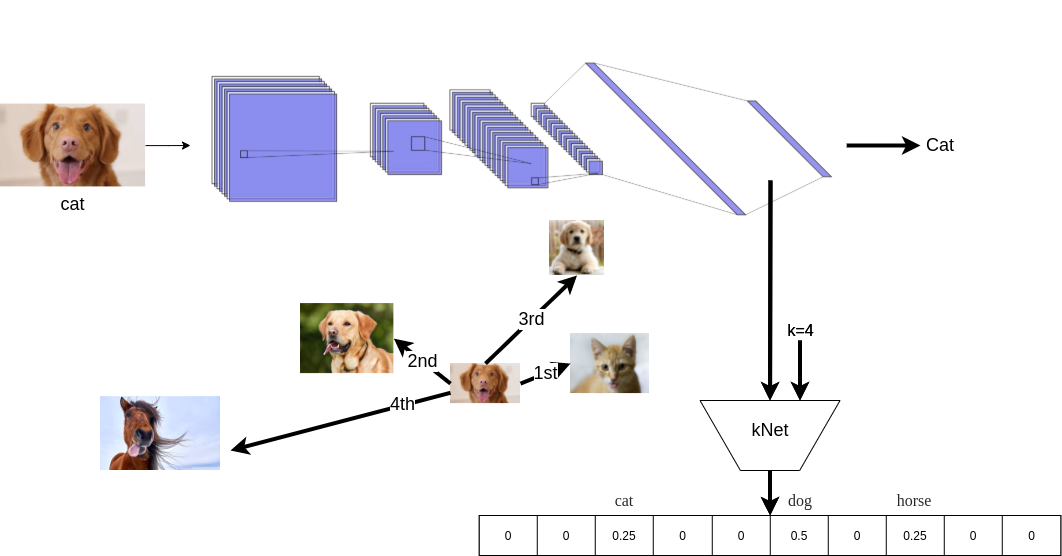} \\
\end{tabular}\\
\caption{\textbf{kNet}. A preliminary network trains on an image of a dog that is wrongly labeled as a "cat". The preliminary network predicts "cat". A kNet network is trained to approximate the output of a kNN classifier of the image, based on the penultimate layer of the preliminary network. Since two of the nearest neighbors are correctly labeled as "dog", kNet outputs "dog" as the label. }
\label{fig:teaser}
\end{center}
\end{figure}

In this work, we train a network, termed kNet, to approximate kNN classification. See Figure~\ref{fig:teaser}. During training the network takes as input a sample point as well as the required $k$ and outputs the distribution of labels, as a kNN algorithm would do. Supplying $k$ as part of the input allows the network to handle different $k$ values at inference time. At inference time, we simply plug the query sample, together with the required $k$ to kNet and obtain the result as if using regular kNN. 

kNet can be trained directly on the raw input data, but we found it beneficial to first train a preliminary network and use the penultimate layer of that network as the representation in which to train kNet.

Approximating kNN is challenging because there are sharp discontinuities in its output, especially when $k$ is small. Approximating these discontinuities requires quite a large network. We use a fairly small model for kNet that approximates kNN well mainly in the large $k$ regime. Experiments show that kNN works well when $k$ is not very small. As a result, kNet serves as a good approximation to kNN in practice.

We conduct experiments on two popular data sets using various types of label noise, different noise levels, and different preliminary networks. In all cases, we observe that kNet improves results consistently and considerably.

% of a given trained network  ( ) on its penultimate layer, which usually being referred as the embedding representation space.
% We showed that even when choosing a trained network that expertise in noise resistance (Zhang et al) or a straight forward cross-entropy approach, applying kNN or our DkNN to the embedding space will enhance the network performance.

% The kNN procedure is highly expensive in terms of memory and run-time.
% Our DkNN network is much cheaper by being added easily as the new head of the given network (replacing the last fully connected layer).

% The nearest neighbor [] classifier has been most of the appealing non-parametric approaches in machine learning. 
% Theoretical []  and empirical []  results show that, for different type of label noise,  k-nearest neighbor (kNN) is robust enough to classify most data correctly.
% Moreover [] show correlation between a DNN's prediction 
% to the labels of the training examples in its local neighborhood.

\section{Related Work}

% The process of collecting reliable labels is becoming more and more expensive. The noise-labels are commonly being contributed to the manual process that might leads to human errors and asymmetric noise (i.e. bird being labeled as an airplane).
% As weak/semi-supervised gaining momentum, pseudo-labels are being generated by a machine and might contains noise labels as well.
% In both cases we do not know which samples are
% incorrectly labeled, neither the total number of mislabeled samples.
%The noisy-label classification task has been studied for long \cite{abs-2007-08199}. 
As the size of labeled data sets grows, so is the concern that some of these labels are noisy. This problem has been investigated extensively, and a survey by Song {\em et al.}~\cite{abs-2007-08199} categorized seven methodological approaches, and provided a comprehensive review of $47$ state-of-the-art studies. The literature on the subject is clear huge, so here we focus on recent research relevant to our work.

\paragraph{Label Noise Modeling}

The most straight forward approach is to remove or re-weight suspected noisy samples by analyzing and searching for inconsistency.

Jeatrakul {\em et al.}~\cite{DBLP:journals/jaciii/JeatrakulWF10} suggested to train a Truth NN and a Falsity NN and eliminate suspected patterns detected by both classifiers.
Wang {\em et al.}~\cite{DBLP:conf/cvpr/WangLMBZSX18} used a Siamese network that minimize the distance between similar samples and maximize distance between dissimilar samples, and re-weight each sample by its likelihood of having a noisy label.   

Predicting the maximum likelihood of a sample label being corrupted has been widely studied as well. A common approach is to model the confusion matrix~\cite{DBLP:conf/iclr/KhetanLA18, DBLP:conf/cvpr/XiaoXYHW15}. Sukhbaatar and Fergus~\cite{SukhbaatarF14}, for example, added a constrained linear layer at the top of the softmax layer, that functions as the transition matrix between the true and the observed noisy labels.

Inspired by them, Goldberger and Ben-Reuven \cite{DBLP:conf/iclr/GoldbergerB17}, optimize the same likelihood function of an EM algorithm that finds the parameters of both the network and the noise.
Wang {\em et al.}~\cite{DBLP:journals/corr/abs-2009-14757} extended it and proposed a Noise-Attention model implemented with multiple noise-specific units to cater to different noisy distributions. 

Another approach is to use kNN and its variants \cite{DBLP:journals/corr/GaoNZ16, DBLP:conf/icitcs/AgrawalR15,DBLP:journals/ml/WilsonM00}. Okamoto and Satoh
~\cite{DBLP:conf/iccbr/OkamotoS95} were the first to provide an expression for the kNN accuracy as a function of noise level. For example, Prasath {\em et al.}~\cite{DBLP:journals/corr/abs-1708-04321} have shown how kNN is resistant to uniform label noise, while Tomasev {\em et al.}~\cite{DBLP:journals/ijon/TomasevB15} proposed several soft weighted voting scheme for the kNN classification, all based on hubness, which express fuzziness of elements appearing in k-neighborhoods of other points. They suggest that such hubness-based fuzzy kNN classification might be suitable for learning under label noise in intrinsically high-dimensional data. 

Recently, Bahri {\em et al.}~\cite{DBLP:conf/icml/BahriJG20} showed how kNN executed on an intermediate layer of a preliminary deep model can help. Specifically, it filters suspiciously-labeled examples and works as well or better than state-of-the-art methods for handling noisy labels. They further show that it is robust to the choice of $k$. 

Such methods are often challenged by the difficulty of distinguishing samples, thus relabeling or removing mislabeled instances may accumulate noise.

\paragraph{Noise-Robust Training}

Minimizing a loss on clean data can be equivalent to the minimization of an approximate loss function on noisy data. Several methods update any surrogate loss to a noise-robust loss
(Patrini {\em et al.}~\cite{DBLP:conf/cvpr/PatriniRMNQ17}, Hendrycks {\em et al.}~\cite{DBLP:conf/nips/HendrycksMWG18},Zhang and Sabuncu~\cite{DBLP:conf/nips/ZhangS18} Lyu and Tsang~\cite{DBLP:conf/iclr/LyuT20}).
Wang~{\em et al.} \cite{DBLP:conf/iccv/0001MCLY019} proposed a symmetric cross entropy loss,
while Thulasidasan {\em et al.}~\cite{DBLP:conf/icml/ThulasidasanBBC19}, updated the cross-entropy loss with an additional output term which is meant to indicate the probability of abstention. 

Meta learners are especially popular in semi-supervised or unsupervised learning settings where the learner generates pseudo labels. Inspired by it, the work of Zhang {\em et al.}~\cite{Zhang20} shows significant improvement in label noise handling and we use it as a preliminary network in our experiments.

\paragraph{DNN vs kNN classification}

The connection between kNN and Deep Neural Network (DNN) has also been investigated. Drory {\em et al.}~\cite{DroryRAG20} showed that the relation between DNN and kNN is especially evident in their performance when trained with noisy data. They observed that DNN robustness to label noise depends on the concentration of noise in the training set. That is, the plurality label (i.e., most common label) in a neighborhood determines the output of the network.

Wang {\em et al.}~\cite{DBLP:conf/icml/WangJC18} showed that the robustness region of kNN approaches that of the Bayes Optimal classifier for fast-growing $k$.
The work of Zhang {\em et al.}~\cite{DBLP:conf/iclr/ZhangBHRV17} demonstrates how DNN is rich enough to memorize the training data and yet generalize pretty well.
Inspired by this, Cohen {\em et al.}~\cite{DBLP:journals/corr/abs-1805-06822} observed that the DNN approximates nearest neighbors decision in embedding space, and such kNN behavior leads to generalization. These works encouraged us to apply kNN on a "noisy" feature space of a trained neural network. 

\paragraph{Soft Labeling}

Soft labeling was first introduced by Szegedy {\em et al.}~\cite{DBLP:conf/cvpr/SzegedyVISW16} as a strategy to improve the performance of a DNN. Rather than standard training with one-hot training labels, a distorted training data distribution is generated, by mixing a uniform label vector with one-hot vectors.

He {\em et al.}~\cite{DBLP:conf/cvpr/HeZ0ZXL19} demonstrate how soft labeling can help 
in terms of generalization.
The work of Muller {\em et al.}~\cite{DBLP:conf/nips/MullerKH19} showed how soft labeling improves model calibration, and encourages the representations of training examples
from the same class to group in tight clusters.
In addition, Lukasik {\em et al.}~\cite{DBLP:conf/icml/LukasikBMK20} have shown that label smoothing is competitive with loss-correction under label noise. Algan and Ulusoy~\cite{DBLP:journals/corr/abs-2007-05836} proposed a meta soft label approach for noisy labels, where in each training iteration a classifier is trained on estimated soft meta labels.

\begin{figure*}[h!]
\begin{center}
\begin{tabular}{c c c c}
\includegraphics[width=0.2\linewidth]{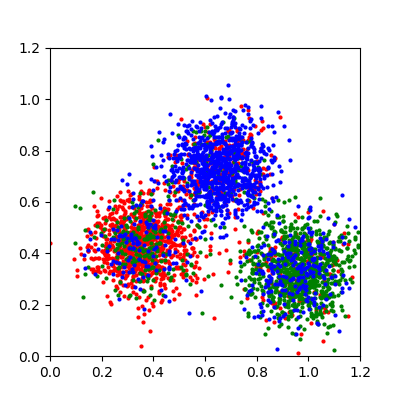} &
\includegraphics[width=0.2\linewidth]{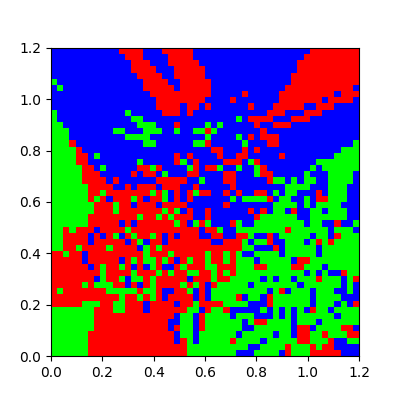} &
\includegraphics[width=0.2\linewidth]{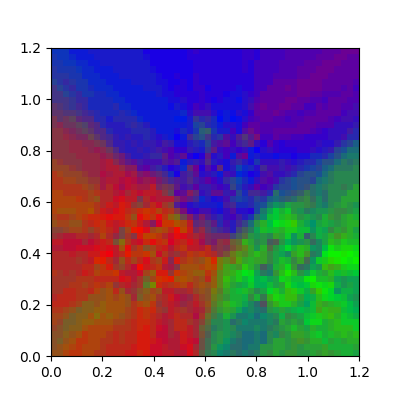} &
\includegraphics[width=0.2\linewidth]{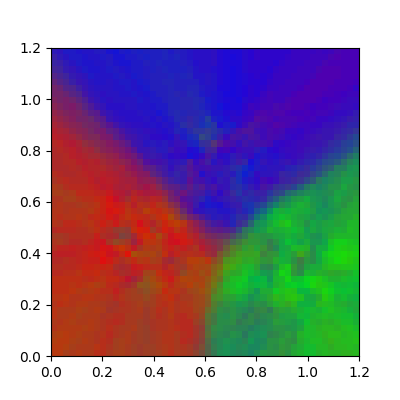} \\
(a) Noisy labels & (b) kNN $(k=1)$ & (c) kNN $(k=19)$ & (d) kNN $(k=49)$ \\
\includegraphics[width=0.2\linewidth]{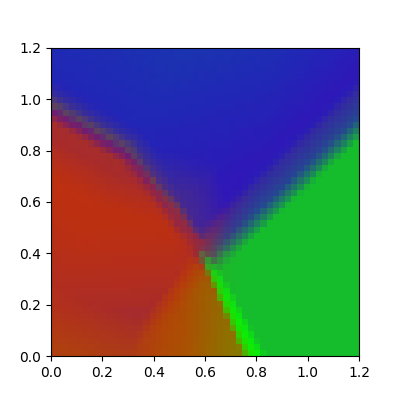} &
\includegraphics[width=0.2\linewidth]{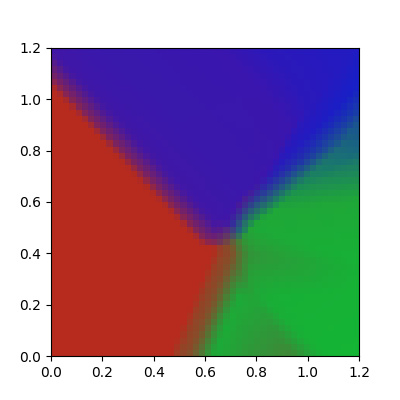} &
\includegraphics[width=0.2\linewidth]{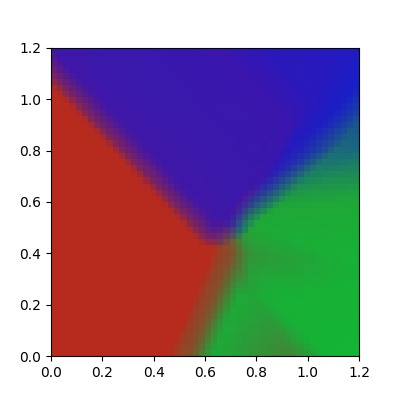} &
\includegraphics[width=0.2\linewidth]{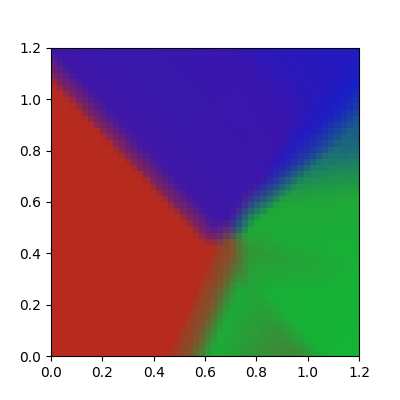} \\
(e) Classification network & (f) kNet $(k=1)$ & (g) kNet $(k=19)$ & (h) kNet $(k=49)$ \\
\end{tabular}\\
~\\
\caption{\textbf{Noisy Labels Toy Example:} (a) noisy samples from 3 classes represented as 3 Gaussians. We add random asymmetric noise by flipping $20\%$ of each class label to the next class and $10\%$ to the third. (e) Classification result of a standard classification network. (b-d) classification results using k-NN with various values of $k$. (f-h) classification results using kNet with various values of $k$. kNN and kNet use the penultimate layer of the classification network (shown in (e)) as their representation. Our work was highly motivated by these results - Unlike the Classification net, kNet predicted a smooth and more accurate boundaries between the classes. In addition, unlike kNN, the inner class prediction of kNet is much more consistent and noise-robust.}
\label{fig:toy_example}
\end{center}
\end{figure*}

% \caption{\textbf{Noisy Labels Toy Example:} (Left column) Top: noisy samples from 3 classes represented as 3 Gaussians. We add asymmetric noise by flipping $20\%$ of each class label to the next class and $10\%$ to the third. Bottom: Clasification result of a standard classification network. (Top) remaining three sub-figures show classification results using k-NN. (Bottom) remaining three sub-figures show classification results using kNet. Our work was highly motivated by these results - Unlike the Classification net, kNet predicted a smooth and more accurate boundaries between the classes. In addition, unlike kNN, the inner class prediction of kNet is much more consistent and noise-robust.}

\section{Method}

We study the standard multi-class classification problem. Given a training set ${\cal T}=\{x_i,y_i\}_{i=1}^n$, where $x$ is the input (i.e., an image) and $y$ is a label in the range ${\cal L}=\{1,...,L\}$, our goal is to label new query point $x$.

We assume label noise, so instead of having ${\cal T}$ we have ${\cal \Tilde{T}}=\{x_i,\Tilde{y}_i\}_{i=1}^n$, where the noisy label $\Tilde{y}$ is sampled from the conditional probability $P(\Tilde{y} | y)$ that can be conveniently modeled by a confusion matrix.

In its simplest form, our approach can be understood as creating a nearest
neighbors classifier in the space defined by the penultimate layer of a trained DNN. 
We first train a standard network on the problem. The penultimate layer of that network gives the representation of the data $x$. Then, we run the kNN algorithm on this representation. It was already observed \cite{DroryRAG20}, that this simple approach can handle label noise quite well.

However, kNN suffers from two shortcomings. First, it has a large memory footprint because it is required to store all the training samples in memory. Second, retrieving the kNN takes time and one often resorts to fast approximate nearest neighbor (ANN) methods~\cite{Arya1998ANNLF} to mitigate this.

Our kNet is a small deep fully connected network that approximates kNN (see architecture in Table~\ref{tab:kNetArchitecture}). The network takes as input the query point $x$ and the desired $k$ and predicts the output of a kNN algorithm. During training we pick a random $k$ and generate a normalized vector $y$ based on the labels of the $k$ nearest neighbours in the embedding space.

The vector $y$ is the main difference between the loss function of the preliminary network and kNet. The preliminary network uses a one-hot vector for $y$, while kNet uses $y$ to represent the distribution of labels in a neighborhood.

% Unlike common classification frameworks we don't necessarily encourage the network to output the correct label, that might lead to wrong classification in case of a noise label set.
% Another way to interpret this is that we encourage kNet to cluster the ambiguously-labeled feature space rather than classify the data.

Motivated by the fact that our network outputs a probability distribution function (the normalized voting vector), we first used KL-divergence loss but found out that in practice, cross-entropy loss converged faster, without accuracy drop.

Replacing kNN with kNet offers a couple of advantages. First, the memory foot print is drastically smaller than that of kNN because there is no need to store the training set. Second, there is no need for Approximate Nearest Neighbor algorithms and the running time is fast, and deterministic.

\begin{table}
\begin{center}
\begin{tabular}{|c|c|c|}
\hline
Type & Filter Shape & Input Size \\ \hline
FC & $d+1 \times d/16$ & $d+1$ \\ \hline
RELU & - & $d/16$ \\ \hline
BN & - & $d/16$ \\ \hline
FC & $d/16 \times L$ & $d/16$ \\ \hline
SOFTMAX & - & $L$ \\ \hline
\end{tabular}
\end{center}
\caption{kNet architecture. $d$ is the dimension of the input sample $x$, and $L$ is the number of labels that $y$ can take.}
\label{tab:kNetArchitecture}
\end{table}

We synthetically generated 2-dimensional points of 3 classes (represented as 3 Gaussians). Next, we added $30\%$ random asymmetric noise which means that each class has $20\%$ chance of being flipped to the first out of the two other classes, $10\%$ chance to flip to the second, and $70\%$ chance to stay the same (e.g. green has $20\%$ to be flipped to blue and $10\%$ to red).  
See Figure~\ref{fig:toy_example}. Panel (a) shows the noisy input, while panel (e) shows the results of a small preliminary network with fully connected layers that was trained using cross entropy loss on the noisy data. 

It seems the classification network overcomes the label noise but it still suffers from non-smooth areas, especially on the border between classes. In addition, the boundary between the red and green classes is not in the right place.

We then use the features from the penultimate layer of the classification network as input to kNN and kNet. See results in panels (b-d) and (f-g) for kNN and kNet, respectively, each with different $k$ values. As can be seen, kNN with $k=1$ is very chaotic and does not handle the label noise well. It improves with larger $k$. In essence, kNN with small $k$'s suffers from non-smooth boundaries while larger $k$'s might lead to wrong classification. 

kNet, on the other hand, gives pretty consistent results for all $k$ values. This shows that the approximation level of kNet is mainly tailored towards larger $k$, because the results there are smoother.

\section{Experiments}

We evaluated our algorithm on the cifar-10 and SVHN datasets, using different preliminary network architectures, different noise models, and different noise levels.

In all cases, kNN was applied to the penultimate layer of the preliminary network, and $k$ was randomly chosen in a range of $[1-101]$. The memory footprint of kNN is determined by the size of the feature vector of the penultimate layer multiplied by the number of samples. In contrast, our kNet is a simple fully-connected based network with only one hidden layer. 
% The motivation of picking a small network was that we want a smooth and even underfit convergence in some noisy samples. 

\paragraph{Label noise}

We considered three noise models in the following experiments. In all cases, the noise was added by a transition matrix specifying the probability of one label being flipped to another. We corrupt the labels in the training data set for a fraction of the examples, the experiment’s “noise rate”, using one of three schemes:

\begin{itemize}
\item {\bf Uniform noise:} The label is flipped to any one of the labels (including itself) with equal probability (the transition matrix is symmetric across the classes)
\item {\bf Semantic asymmetric noise:} A flipped label would be assigned to 
 semantically-similar class (e.g., truck and automobile, or bird and airplane), as was suggested by Patrini {\em et al.}~\cite{DBLP:conf/cvpr/PatriniRMNQ17}.
  \item{\bf Random asymmetric noise:} A flipped label would be assigned to one of two other random labels with probability of 66.6\% to one class and 33.3\% to the second.
 \end{itemize}
 
\paragraph{Fixed kNet} Our toy experiment and the work of \cite{DBLP:conf/icml/BahriJG20} demonstrate how kNet is stable and robust to the choice of $k$. While kNN's accuracy drops dramatically as $k$ gets lower, kNet performance might even get better.
For comparison, we added a fixed kNet experiment where instead of choosing $k$ at random during training, our kNet gets as an input a fixed $k$ in all iterations of training.
Surprisingly (mostly when k=1), the consistency and the performance of kNet was maintained. In some of the fixed kNet experiments, applying it with $k=1$ even gave the best accuracy.

\begin{figure*}[h!]
\begin{center}
\begin{tabular}{c c c}
Toy Example & Cifar10 & SVHN \\ \hline
\includegraphics[width=5cm,height=4cm]{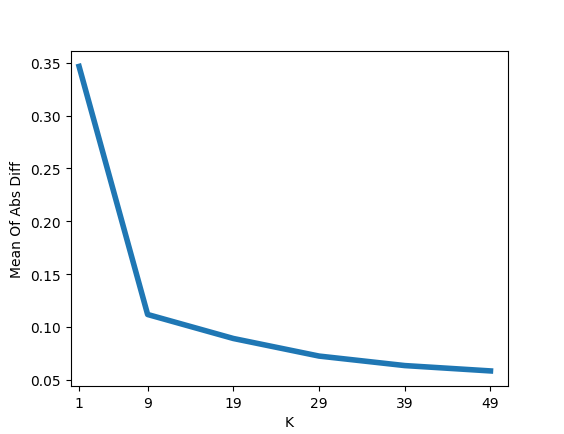} &
\includegraphics[width=5cm,height=4cm]{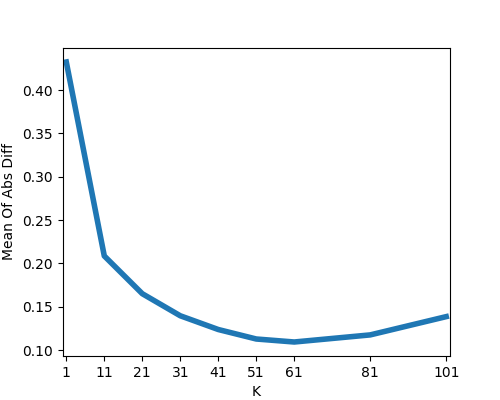} &
\includegraphics[width=5cm,height=4cm]{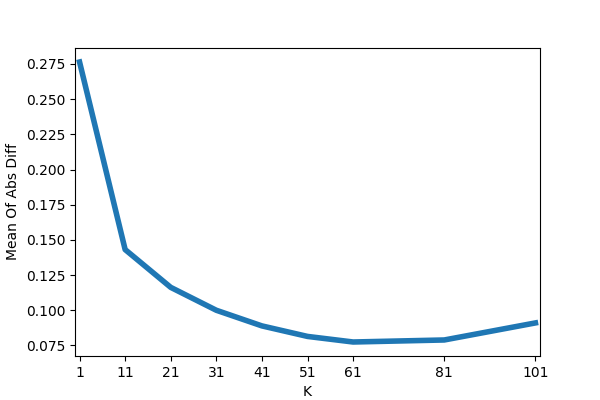} \\
\includegraphics[width=5cm,height=4cm]{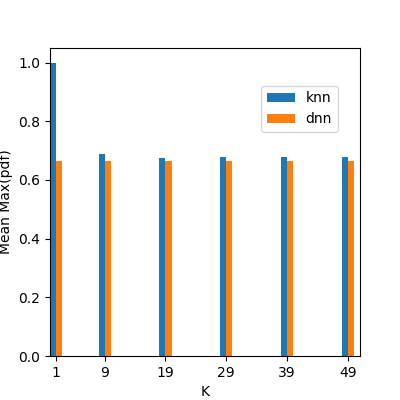} &
\includegraphics[width=5cm,height=4cm]{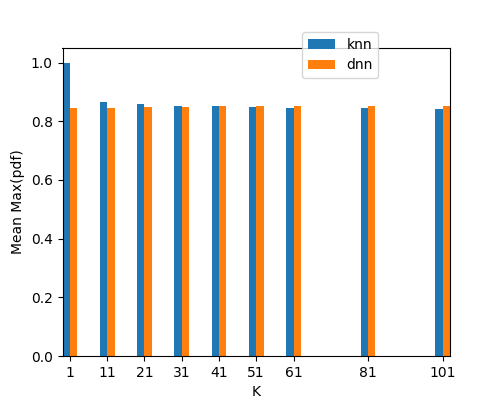} &
\includegraphics[width=5cm,height=4cm]{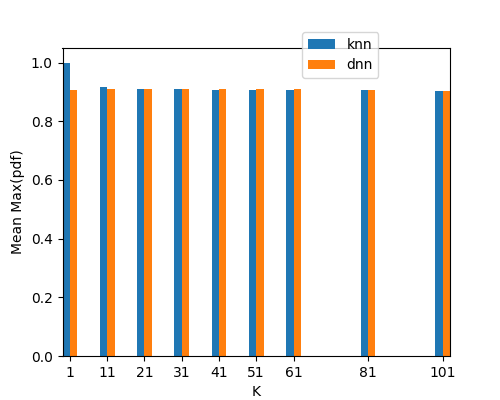} \\
\end{tabular}\\
\caption{\textbf{kNet vs kNN}. A closer look at the correlation between kNet and Knn based on 3 experiments: Toy Example with $30\%$ random asymmetric noise, cifar-10 (Zhang {\em et al.} $80\%$ uniform noise),and SVHN  (Zhang {\em et al.}, $80\%$ uniform noise). (Top) each graph represents the mean absolute difference between the normalized kNN and the kNet prediction for different values of $k$. (bottom) each graph represents the average maximum values of the the normalized kNN and the kNet prediction. From the top row we conclude that kNet best approximates kNN for fairly large $k$ values. From the bottom row we conclude that the maximum output of kNet and kNN is similar, except for very low values of $k$.}
\label{fig:dnn_kk}
\end{center}
\end{figure*}

\paragraph{Preliminary Networks} We consider three preliminary networks in our experiments. The first is based on standard Resnet20. The second is based on a standard 4-depth CNN implementation 
\footnote{We follow \href{https://github.com/tohinz/SVHN-Classifier/blob/master/svhn_classifier.py}{this implementation} which contains 0.55M parameters.},
composed by the standard 2dconv-batchnorm-relu blocks. This simple network has 95.45\% accuracy on the clean SVHN test set.

We have also relied on the much stronger baseline network of Zhang {\em et al.} \cite{Zhang20}. They presented a holistic framework to train deep neural networks in a way that is highly invulnerable to label noise. Their approach re-weights estimated samples that are suspected to have noisy label. In addition, they conduct an ongoing relabeling process based on pseudo labels that are sharp and consistent across augmented versions of input. To the best of our knowledge, their work is the first to demonstrate superior robustness against noise regimes as high as over $90\%$. 

They rely on a prior which is the presence of a small trusted data (i.e. probe data). In our experiments we set $M$, which indicates the number of trusted data used, to be $M=0.01K$. In our case it means 10 images per class.

To train kNet, we first train the three preliminary networks: Resnet20, Zhang {\em et al.}~\cite{Zhang20}, and the CNN mentioned earlier. The size of their penultimate layer is $64, 256$ and $512$ for the three networks, respectively.

Then we trained kNet with a standard cross-entropy (CE) loss, and without adding any noise-robust terms or cleaning data manipulation. The label of each sample when training kNet was given by kNN in the penultiamte layer of the preliminary network. We trained both kNet (where $k$ is part of the input during training) and fixed kNet (where $k$ is fixed during training).

\subsection{End-to-end training with noisy labels}

% Motivated by the toy example, we started by adding a 30\% random asymmetric noise ratio to both cifar-10 and SVHN.
% Each network was trained from scratch, for cifar10 we used a Resnet20, while for SVHN training was done with a standard CNN \footnote{We follow \href{https://github.com/tohinz/SVHN-Classifier/blob/master/svhn_classifier.py}{this implementation} which contains 0.55M parameters}. For both of the experiments the training was done with a standard cross-entropy (CE) loss, without adding any noise-robust terms or cleaning data manipulation. Then, we applied kNN to the penultimate representation of the Resnet20 network and trained a kNet (where $k$ is part of the input during training) and kNet (where $k$ is fixed during training).

\begin{table}
\begin{center}
\scriptsize
\begin{tabular}{|l|c|c|c|c|c|}
\hline
\multicolumn{6}{|c|}{ {\bf cifar-10} }\\
\hline
Network & \# Parameters & \multicolumn{4}{c|}{ K } \\ \hline
& & 1 & 11 & 51 & 101 \\ \hline
Resnet20 & 0.27M & \multicolumn{4}{c|}{ 66.43 } \\ \hline
~~~ + kNN & 0.27M + 3.2M & 38.95 & 62.01 & 65.72 & {\bf 66.92} \\ \hline
~~~ + kNet & 0.27M + 3K & 67.05 & {\bf 67.16} & 66.93 & 65.99 \\ \hline 
Zhang{\em et al.} \cite{Zhang20} & 0.87M & \multicolumn{4}{c|}{ 89.58 } \\ \hline
~~~ + kNN & 0.87M + 12.8M & 86.92 & {\bf 89.56} & 89.42 & 89.27 \\ \hline
~~~ + kNet & 0.87M + 5K & 89.24 & 89.43 & {\bf 89.63} & 89.53 \\ \hline
~~~ + fixed kNet & 0.87M + 5K & \textbf{89.88} & 89.84 & 89.34 & 89.2 \\ \hline
\multicolumn{6}{|c|}{ {\bf SVHN} }\\
\hline
Network & \# Parameters & \multicolumn{4}{c|}{ K } \\ \hline
& & 1 & 21 & 61 & 101 \\ \hline
CNN & 0.55M & \multicolumn{4}{c|}{ 92.92 } \\ \hline
~~~ + kNN & 0.55M+37.5M &  56.45 & 92.13 & 93.03 & {\bf 93.07}  \\ \hline
~~~ + kNet & 0.55M+4K & 92.97 & 92.99 & 93.01 & {\bf 93.02} \\ \hline
Zhang {\em et al.} \cite{Zhang20} & 0.87M & \multicolumn{4}{c|}{ 91.88 } \\ \hline
~~~ + kNN & 0.87M + 12.8M & 90.14 & \textbf{93.39} & 93.2 & 93.07 \\ \hline
~~~ + kNet & 0.87M + 5K & 93.09 & 93.07 & {\bf 93.11} & 93.08 \\ \hline
~~~ + fixed kNet & 0.87M + 5K & 92.88 & {\bf 92.9} & 92.81 & 92.73 \\ \hline
\end{tabular}
\end{center}
\caption{ {\bf ${\bf 40\%}$ uniform label noise:} We show results on both cifar-10 (top) and SVHN (bottom) when adding $40\%$ of uniform label noise. kNN, kNet, and fixed kNet run on the penultimate layer of different preliminary networks: ResnNet20, CNN, and Zhang {\em et al.}~\cite{Zhang20}. Best result in each row is marked in {\bf bold}. kNet and fixed kNet always improve the preliminary network. kNet requires $3K$ to $5K$ parameters, compared to $3.2M$ up to $12.8M$ parameters of kNN.}
\label{tab:uniform_40}
\end{table}

\begin{table}
\begin{center}
\scriptsize
\begin{tabular}{|l|c|c|c|c|c|}
\hline
\multicolumn{6}{|c|}{ {\bf cifar-10} }\\
\hline
Network & \# Parameters & \multicolumn{4}{c|}{ K } \\ \hline
& & 1 & 11 & 51 & 101 \\ \hline
Resnet20 & 0.27M & \multicolumn{4}{c}{ 23.71 } \\ \hline
~~~ + Knn & 0.27M + 3.2M & 12.85 & 14.66 & 14.66 & {\bf 22.27} \\ \hline
~~~ + kNet & 0.27M + 5K & 27.9 & 27.93 & {\bf 27.93} & 27.17 \\ \hline
Zhang {\em et al.} \cite{Zhang20} & 0.87M & \multicolumn{4}{c|}{ 84.87 } \\ \hline
~~~ + kNN & 0.87M + 12.8M & 82.41 & 86.11 & \textbf{86.16} & 85.85 \\ \hline
~~~ + kNet & 0.87M + 5K & 84.81 & 85.35 & {\bf 85.5} & 85.25 \\ \hline
~~~ + fixed kNet & 0.87M + 5K & {\bf 85.99} & 85.89 & 85.38 & 85.27 \\ \hline
\multicolumn{6}{|c|}{ {\bf SVHN} }\\
\hline
Network & \# Parameters & \multicolumn{4}{c|}{ K } \\ \hline
& & 1 & 11 & 41 & 101 \\ \hline
CNN & 0.27M & \multicolumn{4}{c}{ 76.57 } \\ \hline
~~~ + kNN & 0.27M + 37.5M & 22.21 & 43.68 & 66.63 & {\bf 74.15} \\ \hline
~~~ + kNet & 0.27M + 5K & 76.92 & 76.96 & {\bf 77.09} & 77.01 \\ \hline
Zhang {\em et al.} \cite{Zhang20} & 0.87M & \multicolumn{4}{c|}{ 89.26 } \\ \hline
~~~ + kNN & 0.87M + 12.8M & 88.28 & 91.69 & {\bf 91.7} & 91.61 \\ \hline
~~~ + kNet & 0.87M + 5K & 91.34 & 91.46 & {\bf 91.57} & 91.31 \\ \hline
~~~ + fixed kNet & 0.87M + 5K & 91.43 & \textbf{91.81} & 91.42 & 91.09 \\ \hline
\end{tabular}
\end{center}
\caption{ {\bf ${\bf 80\%}$ uniform label noise:} We show results on both cifar-10 (top) and SVHN (bottom) when adding $80\%$ of uniform label noise. kNN, kNet, and fixed kNet run on the penultimate layer of different preliminary networks: ResnNet20, CNN, and Zhang {\em et al.}~\cite{Zhang20}. Both ResNet20 and CNN fail catastrophically in this case, and the kNN variants do not save the day. However kNet improves Zhang {\em et al.}~\cite{Zhang20} on both data sets by up to $2\%$. Best result in each row is marked in {\bf bold}. kNet and fixed kNet always improve the preliminary network. kNet requires $3K$ to $5K$ parameters, compared to $3.2M$ up to $12.8M$ parameters of kNN.}
\label{tab:uniform_80}
\end{table}

\paragraph{Uniform Label Noise:}

In the first experiment, we add $40\%$ uniform label noise and report the results for cifar-10 and SVHN in Table~\ref{tab:uniform_40}. We see that Resnet20 behaves poorly in this experiment. kNN (except for the kNN with $k=1$ variant that fails miserably) as well as kNet and fixed kNet improve the preliminary networks for some values of $k$.

A similar pattern can be observed when using the much stronger baseline network of Zhan {\em et al.} \cite{Zhang20}. Our kNN variants improve their performance by up to $2\%$ in the case of cifar-10 and $0.3\%$ in the SVHN case. In both cases, the kNet network requires a modest increase of $3K$ to $5K$ in memory. The simple CNN network performs almost as well as the method of Zhang {\em et al.}~\cite{Zhang20}, yet kNet variants improve it slightly. 

We repeated this experiment with $80\%$ uniform noise (Table~\ref{tab:uniform_80}). This time, the CNN network fails catastrophically on the SVHN data set, while the method of Zhang {\em et al.}~\cite{Zhang20} performs well. We note that a kNN variant improves the baseline network by about $2\%$ in both cifar-10 and SVHN. 

kNet and fixed kNet are quite similar across a range of $k$ values. This indicates that the networks are not sensitive to $k$, as opposed to the classic kNN classifier that behaves differently for different values of $k$.

\begin{table}
\begin{center}
\scriptsize
\begin{tabular}{|l|c|c|c|c|c|c|}
\hline
\multicolumn{6}{|c|}{ {\bf cifar-10} }\\
\hline
Network & \# Parameters & \multicolumn{4}{c|}{ K } \\ \hline
& & 1 & 11 & 61 & 101 \\ \hline
Resnet20 & 0.27M & \multicolumn{4}{c|}{ 75.28 } \\ \hline
~~~ + kNN & 0.27M+3.2M & 61.49 & 85.51 & {\bf 87.74} & 87.56 \\ \hline
~~~ + kNet & 0.27+3K & 87.49 & {\bf 87.53} & 87.38 & 87.38 \\ \hline
Zhang {\em et al.}~\cite{Zhang20} & 0.87M & \multicolumn{4}{c|}{ 88.79 } \\ \hline
~~~ + kNN & 0.87M+12.8M & 85.23 & 88.4 & {\bf 88.51} & 88.51 \\ \hline
~~~ + kNet & 0.87M+5K & 88.63 & 88.84 & {\bf 88.89} & 88.35 \\ \hline
~~~ + fixed kNet & 0.87M+5K & \textbf{88.94} & 88.88 & 88.91 & 88.87 \\ \hline
\multicolumn{6}{|c|}{ {\bf SVHN} }\\
\hline
Network & \# Parameters & \multicolumn{4}{c|}{ K } \\ \hline
& & 1 & 11 & 51 & 101 \\ \hline
CNN (CE loss) & 0.55M & \multicolumn{4}{c|}{ 88.23 } \\ \hline
~~~ + kNN & 0.55M+37.5M & 61.08 & 85.2 & 89.4 & {\bf 89.56} \\ \hline
~~~ + kNet & 0.55M+4K & 89.27 & 89.27 & {\bf 89.28} & 89.24 \\ \hline
Zhang {\em et al.}~\cite{Zhang20} & 0.87M & \multicolumn{4}{c|}{ 92.06 } \\ \hline
~~~ + kNN & 0.87M+12.8M & 90.56 & {\bf 93.47} & 93.36 & 93.24 \\ \hline
~~~ + kNet & 0.87M+5K & 92.36 & 92.71 & {\bf 93.02} & 92.88 \\ \hline
~~~ + fixed kNet & 0.87M+5K & \textbf{93.4} & 93.04 & 93.12 & 93.07 \\ \hline

\end{tabular}
\end{center}
\caption{ {\bf ${\bf 30\%}$ asymmetric label noise:} We show results on both cifar-10 (top) and SVHN (bottom) when adding $30\%$ of asymmetric label noise. kNN, kNet, and fixed kNet run on the penultimate layer of different preliminary networks: ResnNet20, CNN, and Zhang {\em et al.}~\cite{Zhang20}. Best result in each row is marked in {\bf bold}. kNet and fixed kNet always improve the preliminary network. kNet requires $3K$ to $5K$ parameters, compared to $3.2M$ up to $12.8M$ parameters of kNN.}
\label{tab:asymetric_30}
\end{table}

\paragraph{Asymmetric Label Noise:} Next, we evaluated the case of asymmetric label noise with $30\%$ noise rate. This is a setting similar to the one reported in the toy example.
We performed this experiment for both cifar-10 and SVHN and report results in Table \ref{tab:asymetric_30}. As can be seen, in both cases, all three kNN variants (kNN, kNet, and fixed kNet) improve the performance of the base Resnet20 and CNN networks. In the case of cifar-10, the improvement is quite dramatic (from about $75\%$ to about $88\%$ in all three variants). 

On both data sets, kNN with $k=1$ performs much worst than the base Resnet20 network. Results improve dramatically as we increase $k$. The classification results for both kNet and fixed kNet are quite similar and consistent across the different $k$ values. This is to be expected because our network is quite small and offers a very smooth approximation of kNN, which corresponds to classification with a large $k$. In addition, observe that the memory footprint of kNet and fixed kNet is just $3K$ or $4K$, compared to $3.2M$ and $37.5M$ for kNN, on cifar-10 and SVHN, respectively.

\paragraph{Semantic asymmetric Label Noise:} In the last experiment, we evaluated the case of semantic asymmetric label noise with $40\%$ noise rate. Again, we performed the experiment for both cifar-10 and SVHN and report results in Table \ref{tab:symmetric_40}. As can be seen, in both cases, either kNet or fixed kNet improved the performance of the base Resnet20, CNN and Zhang {\em et al.}~\cite{Zhang20} networks.

\begin{table}
\begin{center}
\scriptsize
\begin{tabular}{|l|c|c|c|c|c|c|}
\hline
\multicolumn{6}{|c|}{ {\bf cifar-10} }\\
\hline
Network & \# Parameters & \multicolumn{4}{c|}{ K } \\ \hline
& & 1 & 21 & 51 & 101 \\ \hline
Resnet20 & 0.27M & \multicolumn{4}{c}{ 70.35 } \\ \hline
~~~ + kNN & 0.27M + 3.2M & 61.29 & 66.53 & 70 & \textbf{70.03} \\ \hline
~~~ + kNet & 0.27M + 3K & 71.45 & 71.51 & \textbf{71.65} & 71.47 \\ \hline
Zhang {\em et al.}~\cite{Zhang20} & 0.87M & \multicolumn{4}{c|}{ 89.76 } \\ \hline
~~~  + kNN & 0.87M + 12.8M & 85.79 & {\bf 89.29} & 89.16 & 88.79 \\ \hline
~~~  + kNet & 0.87M + 5K & 88.94 & 89.12 & {\bf 89.45} & 88.72 \\ \hline
~~~  + fixed kNet & 0.87M + 5K & \textbf{89.91} & 89.58 & 89.77 & 89.63 \\ \hline
\multicolumn{6}{|c|}{ {\bf SVHN} }\\
\hline
Network & \# Parameters & \multicolumn{4}{c|}{ K } \\ \hline
& & 1 & 21 & 41 & 101 \\ \hline
CNN & 0.55M & \multicolumn{4}{c|}{ 93.6 } \\ \hline
~~~ + Knn & 0.55M+37.5M & 73.33 & 88.02 & 89.67 & \textbf{93.18} \\ \hline
~~~ + kNet & 0.55M+4K & 93.85 & 93.97 & 94 & \textbf{94.05} \\ \hline

Zhang {\em et al.}~\cite{Zhang20} & 0.87M & \multicolumn{4}{c|}{ 92.59 } \\ \hline
~~~  + kNN & 0.87M + 12.8M & 91.69 & 94.09 & \textbf{94.11} & 93.99 \\ \hline
~~~  + kNet & 0.87M + 5K & 93.3 & 93.51 & {\bf 93.78} & 93.4 \\ \hline
~~~  + fixed kNet & 0.87M + 5K & 93.83 & 93.83 & {\bf 93.85} & 93.84 \\ \hline
\end{tabular}
\end{center}
\caption{ {\bf ${\bf 40\%}$ semantic asymmetric label noise:} We show results on both cifar-10 (top) and SVHN (bottom) when adding $40\%$ of semantic asymmetric label noise. kNN, kNet, and fixed kNet run on the penultimate layer of different preliminary networks: ResnNet20, CNN, and Zhang {\em et al.}~\cite{Zhang20}. Best result in each row is marked in {\bf bold}. kNet and fixed kNet always improve the preliminary network. kNet requires $3K$ to $5K$ parameters, compared to $3.2M$ up to $12.8M$ parameters of kNN.}

\label{tab:symmetric_40}
\end{table}

\begin{figure*}[h!]
\centering
\begin{tabular}{c c}
\includegraphics[width=0.5\linewidth]{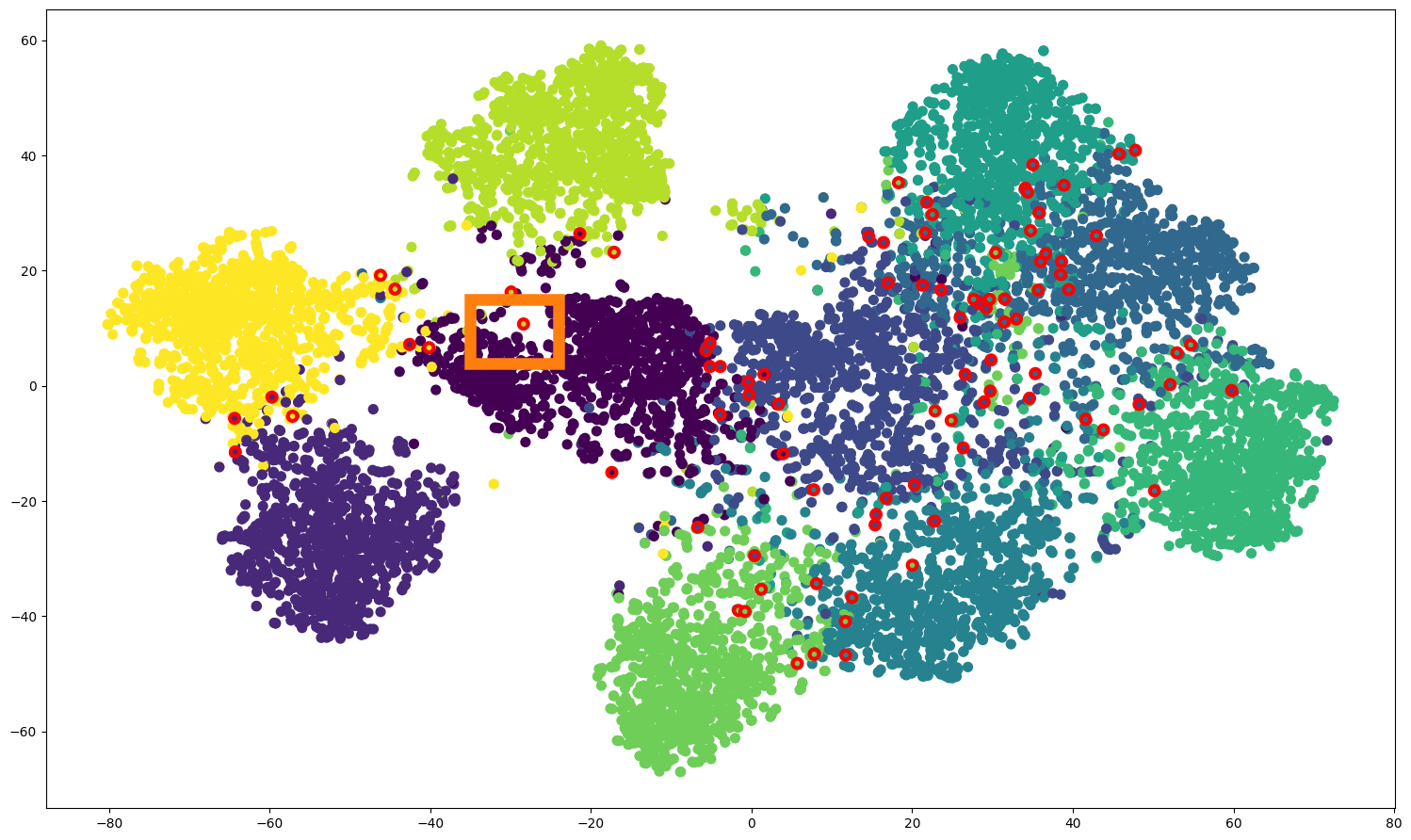} &
\includegraphics[width=0.5\linewidth]{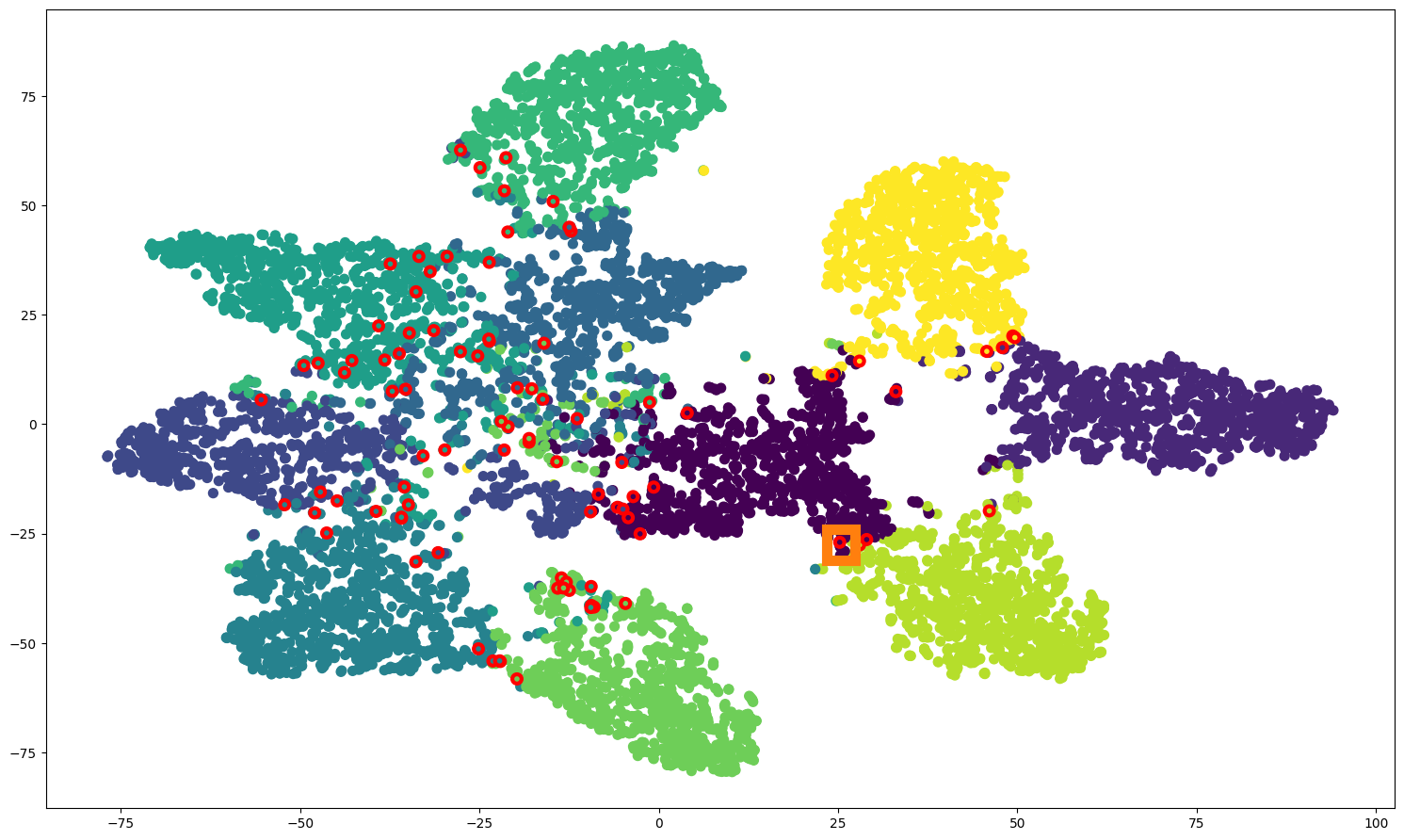} \\
(a) t-SNE: Zhang {\em et al.} & (b) t-SNE:  kNet\\
\includegraphics[width=0.5\linewidth]{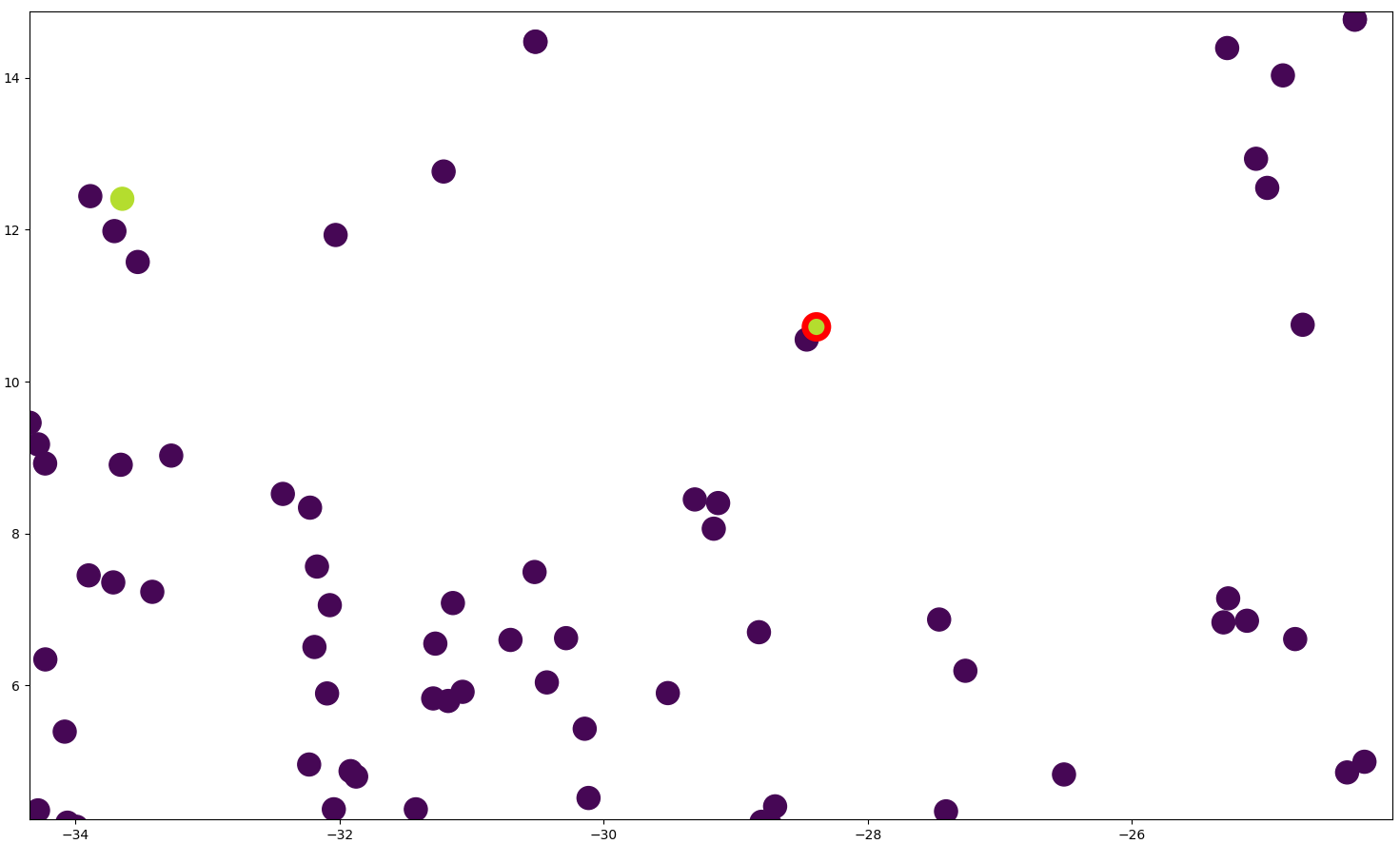} &
\includegraphics[width=0.5\linewidth]{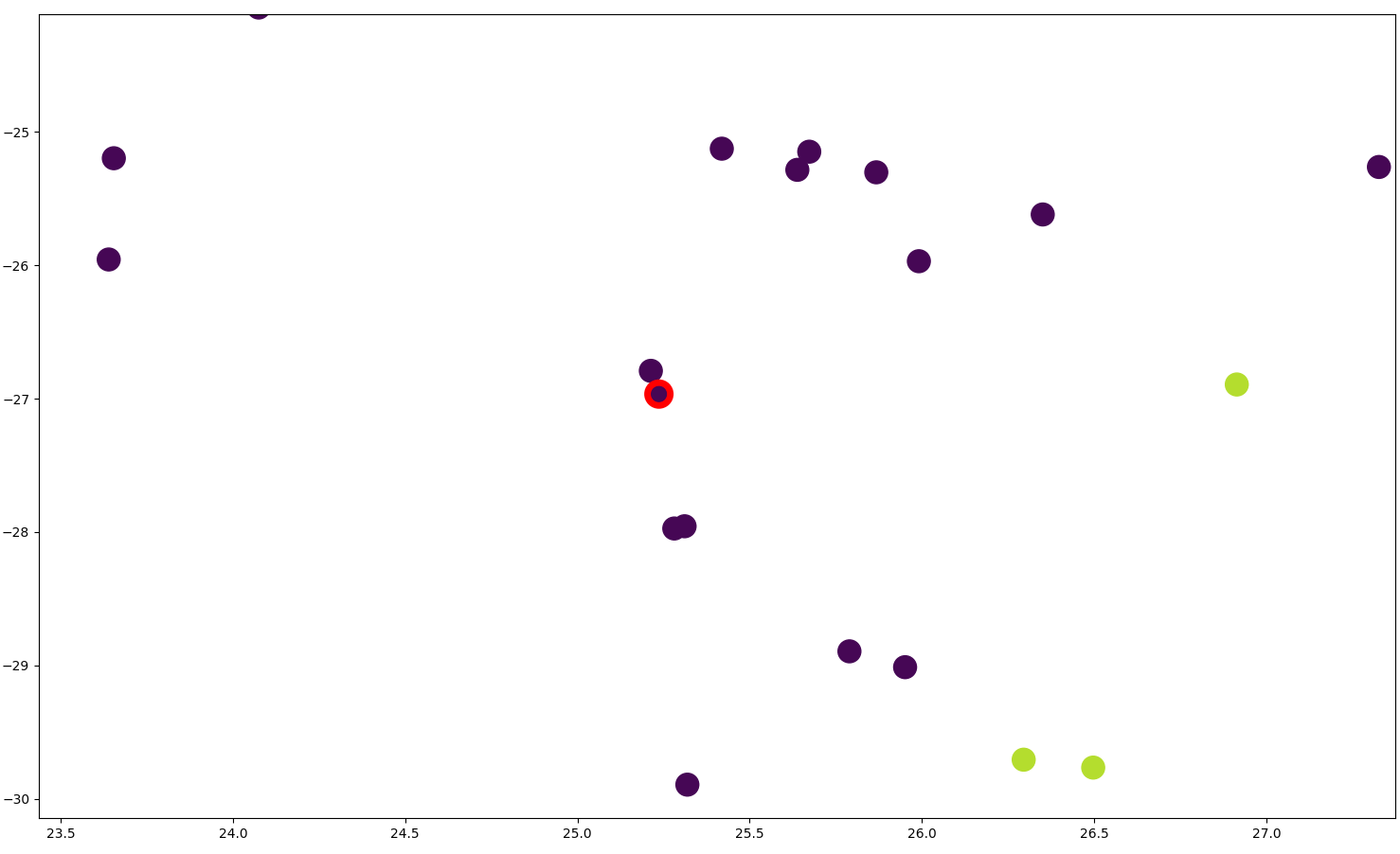} \\
(c) Zoomed-in t-SNE: Zhang {\em et al.}  & (d) Zoomed-in t-SNE: kNet \\
\end{tabular}
\caption{\textbf{t-SNE: Cifar10} (a)-(b): t-SNE on the penultimate layer of Zhang {\em et al.} and kNet respectively. The visualization supports our toy example conclusions - kNet provides a better class separation and concentration (less mixed clusters).
The red circled points represents wrong classified points using Zhang {\em et al.} that were classified correctly using kNet. The regions bounded by the orange boxes are zoomed-in and presented below each figure.
(c)-(d): A zoomed-in illustrations. The purple/green points are car/truck predictions respectively. The two red circled points represents the same query image. (c) The red circled represents a truck image that was wrongly classified as a car. This point could have been classified correctly by applying kNN. (d) The same truck image (represented by the red circled point) is now being classified correctly as a car using kNet.    }
\label{fig:t-sne}
\end{figure*}

\paragraph{kNet Approximation:} As can be seen in the toy example (Figure~\ref{fig:toy_example}), kNet tends to smooth out the results. This is further evident in the experiments where the best performing kNet had a value of $k=51$ or $k=61$. Intuitively, this suggests that kNet approximates kNN with a large $k$. 

To validate this assumption we compared the probability distribution function (pdf) of kNN with the one produced by kNet, for different values of $k$. Results can be seen in Figure~\ref{fig:dnn_kk}. On top we show the mean absolute difference between the pdf of kNet and that of kNN, for various $k$ values. There is a large gap at low $k$ values, and the minimum is achieved at around $50$ or $60$. This observation holds true for all data sets considered in this work.

Figure~\ref{fig:dnn_kk} (bottom) shows the maximum value in the pdf of both kNN and kNet. The maximum value is almost identical for both methods, except for $k=1$ where there is a large gap between the two. We conclude that kNet approximates kNN with a large $k$.

\paragraph{How to choose $k$?}
kNet gives the user the freedom to choose $k$ at inference time. However, it is not clear what value of $k$ to choose. It appears that choosing $k=41$ or $k=15$ usually gives good results. Alternatively, we observed that if we train fixed kNet with $k=1$ we get consistently excellent results. 

\paragraph{t-SNE visualization}
In order to get a visual comparison of kNet and \cite{Zhang20}, we apply a popular dimensionality reduction algorithm,t-distributed Stochastic Neighbor Embedding (t-SNE, \cite{tsne}),
on the penultimate layer of \cite{Zhang20} and our kNet model.
We applied it on the networks that were trained on the Cifar10 with 40\% uniform noise. 
As can be seen in ~\ref{fig:t-sne}, \cite{Zhang20} suffers more from inconsistent boundaries between classes, and wrong classifications(marked with red circles) that could have been correct using kNN.
The penultimate layer of kNet is basically the first layer that projects the feature space 1+256-dimensional of \cite{Zhang20} to 16-dimensional space.
The boundaries between classes are more clear and all of the red circled points that were wrongly classified by \cite{Zhang20} are now correctly classified by kNet.\newline

\section{Conclusions}

Deep Neural Networks must handle label noise. kNN is a viable option to deal with this problem, but kNN requires a large memory footprint, which makes it less desirable in practice. Therefore, we propose kNet.

kNet is a network that approximates kNN and improves the classification results of preliminary networks that were trained with noisy labels. kNet runs on top of the penultimate layer of the preliminary network and we have shown its effectiveness on different data sets (cifar-10 and SVHN), different preliminary networks (Resnet20, CNN, and Zhang {\em et al.} \cite{Zhang20}), different label noise models (uniform, random asymmetric, and semantic asymmetric noise models) and different noise levels ($30\%$, $40\%$, and $80\%$). 

The memory footprint of kNet is about $4K$ parameters, which is orders of magnitude smaller than the entire dataset required by kNN. And since kNet is a neural network, running it at inference time is fast, deterministic, and does not require advanced data structure such as approximate nearest neighbors that are often required in kNN algorithms.

% {\small
% \bibliographystyle{ieee_fullname}
% \bibliography{egbib}
% }

\end{document}